\documentclass[10pt,twocolumn,letterpaper]{article}

\usepackage{cvpr}              
\usepackage{multirow}
\usepackage[dvipsnames]{xcolor}

\definecolor{cvprblue}{rgb}{0.21,0.49,0.74}
\usepackage[pagebackref,breaklinks,colorlinks,citecolor=cvprblue]{hyperref}

\def\paperID{351} 
\def\confName{CVPR}
\def\confYear{2024}

\title{SpikeNeRF: Learning Neural Radiance Fields from Continuous Spike Stream}

\author{Lin Zhu$^1$
\and
Kangmin Jia$^1$
\and
Yifan Zhao$^2$
\and
Yunshan Qi$^2$
\and
Lizhi Wang$^1$
\and
Hua Huang$^{3,1,}$\thanks{Corresponding author: Hua Huang.}
\and
$^1$Beijing Institute of Technology
\and
$^2$Beihang University
\and
$^3$Beijing Normal University
}

\begin{document}

\maketitle

\begin{abstract}
Spike cameras, leveraging spike-based integration sampling and high temporal resolution, offer distinct advantages over standard cameras. However, existing approaches reliant on spike cameras often assume optimal illumination, a condition frequently unmet in real-world scenarios. To address this, we introduce SpikeNeRF, the first work that derives a NeRF-based volumetric scene representation from spike camera data.
Our approach leverages NeRF's multi-view consistency to establish robust self-supervision, effectively eliminating erroneous measurements and uncovering coherent structures within exceedingly noisy input amidst diverse real-world illumination scenarios. 
The framework comprises two core elements: a spike generation model incorporating an integrate-and-fire neuron layer and parameters accounting for non-idealities, such as threshold variation, and a spike rendering loss capable of generalizing across varying illumination conditions.
We describe how to effectively optimize neural radiance fields to render photorealistic novel views from the novel continuous spike stream, demonstrating advantages over other vision sensors in certain scenes.
Empirical evaluations conducted on both real and novel realistically simulated sequences affirm the efficacy of our methodology. The dataset and source code are released at \url{https://github.com/BIT-Vision/SpikeNeRF}.

\end{abstract}    
\section{Introduction}
\label{sec:intro}

In recent years, there has been significant progress in the development of neuromorphic cameras, with notable advancements such as event cameras~\cite{delbruck2010activity, 2} and spike cameras~\cite{recon,huang20231000}. These innovative devices excel in capturing light intensity changes in high-speed scenes, presenting a breakthrough in the realm of visual perception~\cite{zhu2021neuspike,time-lens,etrack1,edepth1,eflow3,gu2023reliable,wang2023visevent}. Spike camera, wherein each pixel responds independently to the accumulation of photons by generating asynchronous spikes~\cite{recon}. This distinctive feature enables the spike camera to record full visual details with an ultra-high temporal resolution of up to 40 kHz (or 20 Hz for the portable version).
These advantages include the capability to capture rapid changes in scenes and efficiently represent dynamic environments. With these unique features, spike cameras have demonstrated superiority in handling multiple computer vision tasks~\cite{zhu2023recurrent,zhu2020hybrid,zhu2021neuspike,wang2022learning,xiang2021learning,zhao2021spk2imgnet,Han_2020_CVPR,hu2022optical,zhu2020retina}.

In parallel, there has been a growing trend in the computer vision community to explore neural radiance fields (NeRFs) as a solution for scene representation and novel view synthesis~\cite{NeRF,Deblur-NeRF,fastnerf,depthnerf,flownerf,pixelnerf,regnerf,enerf1,enerf2}. NeRFs employ a multilayer perceptron (MLP) combined with differentiable rendering to represent scenes, allowing for the synthesis of novel views from unseen perspectives.
So far, NeRFs have predominantly undergone examination using simulated data and high-quality real-world images acquired under optimal conditions. Some studies have explored the application of NeRF to a distinct type of neuromorphic camera known as an event camera, characterized by differential sampling.
Noteworthy among these endeavors are Ev-NeRF \cite{enerf1} and EventNeRF \cite{enerf2}, both of which introduced neural radiance fields derived exclusively from event streams. However, the inherent lack of texture details in event data has constrained the effectiveness of these approaches, resulting in limited outcomes. 
Instead, a spike camera can offer fine texture details through a high temporal resolution spike stream.

\begin{figure*}[t]
  \centering
    \includegraphics[width=1\linewidth]{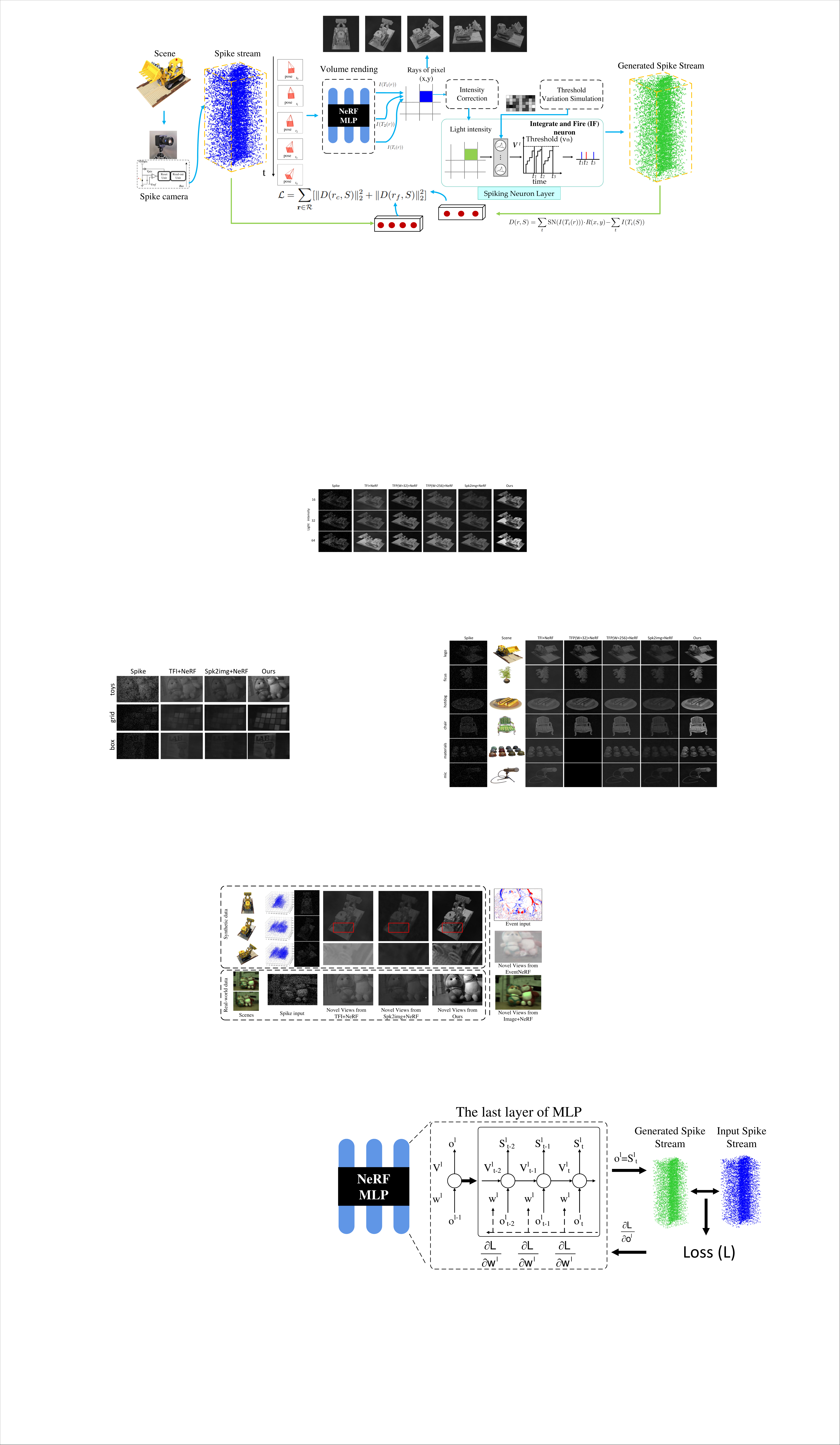}
    \vspace{-6mm}
    \caption{Comparing novel views across various vision sensors, our SpikeNeRF stands out as the first method to learn neural radiance fields from a continuous spike stream. The Spike camera, operating at 20,000 Hz, eliminates motion blur, distinguishing it from traditional cameras. Notably, when contrasted with other methods (e.g., TFI+NeRF~\cite{recon}, Spk2img+NeRF~\cite{zhao2021spk2imgnet}, and EventNeRF~\cite{enerf2}) and sensors (e.g., event camera), the rendered views of objects or scenes exhibit significantly enhanced sharpness.}
   \vspace{-2mm}
   \label{fig:1}
\end{figure*}

However, their application to spike camera data poses a unique challenge due to the distinctive data modality.
The output of a spike camera is a spike stream, fundamentally different from traditional images. Each pixel in a spike camera generates a spike, and the accumulation is reset when the photon accumulation exceeds a predefined threshold. At each timestamp, the spike camera outputs a binary matrix, referred to as a spike frame, indicating the presence of spikes at all pixels. Currently, there is no established spike-based method for addressing the challenge of generating dense, photorealistic 3D representations of scenes. 
Therefore, the challenge tackled in this paper is: \emph{Can we reconstruct a dense volumetric 3D representation of the scene from a spike stream captured by a moving spike camera?}

Addressing the aforementioned challenges, this paper introduces SpikeNeRF, the first approach for deriving a volumetric scene representation from spike camera data using Neural Radiance Fields (NeRF). Our method is designed for purely spike-based supervision, preserving texture and motion details in high temporal resolution. We evaluate the efficacy of SpikeNeRF for the task of novel view synthesis on a new dataset of synthetic and real spike sequences. SpikeNeRF leverages NeRF's inherent multi-view consistency to establish robust self-supervision, mitigating the impact of erroneous measurements in the presence of high noise levels and diverse illumination conditions. 
Specifically, SpikeNeRF incorporates a spike generation model based on a spiking neuron layer, which considers intrinsic parameters and non-idealities associated with spike cameras. 
This model enhances the fidelity of spike data, providing a more accurate representation of the scene under varying lighting conditions (see Fig.~\ref{fig:1}).
To enable generalization across diverse illumination scenarios, SpikeNeRF introduces a long-term spike rendering loss. This tandem of loss functions ensures that the model can effectively capture and represent scene geometry, even in challenging lighting conditions.
The main contributions of this paper are:

1) We present the first approach for inferring a NeRF volume from only a spike stream. SpikeNeRF is highly robust to various illumination conditions and builds a coherent 3D structure that can provide high-quality observations.

2) We develop a long-term spike rendering loss based on a spiking neuron layer and threshold variation simulation, which are effective in enhancing neural volumetric representation learning.

3) We build both synthetic and real-world datasets for training and testing our model. Experimental results demonstrate that our method outperforms existing methods. We release the newly recorded spike dataset and source code to facilitate the research field.


\section{Related Work}
\label{sec:formatting}
\subsection{Neural Radiance Fields}
In recent years, Neural Radiance Fields have garnered significant attention for their remarkable performance in tasks involving neural implicit 3D representation and novel view synthesis. Various enhancements, such as FastNeRF \cite{fastnerf} and Depth-supervised NeRF \cite{depthnerf}, aim to expedite NeRF's learning speed. Neural scene flow fields \cite{flownerf} delve into 3D Scene Representation Learning for dynamic scenes. Approaches like PixelNeRF \cite{pixelnerf} and RegNeRF \cite{regnerf} leverage a minimal number of input images for high-quality novel view synthesis. Mip-NeRF \cite{mipnerf} introduces a frustum-based sampling strategy for NeRF-based anti-aliasing, addressing artifacts while enhancing training speed. Beyond speed and data augmentation, some works focus on enhancing NeRF with low-quality input images. For instance, NeRF in the wild \cite{nerfinw} utilizes low-quality images captured by tourists, tackling occlusions and inconsistent lighting conditions. NeRF in the dark \cite{nerfind} and HDR-NeRF \cite{hdr-nerf} enable the synthesis of high dynamic range new view images from noisy low dynamic images. Additionally, Deblur-NeRF \cite{Deblur-NeRF} introduces the DSK module for simulating the blurring process, facilitating novel view synthesis from blurry to sharp images. However, Deblur-NeRF may face challenges in scenarios where the camera coincidentally shakes uniformly across all views or the input images exhibit strong blur.

\subsection{Scene Reconstruction based on Neuromorphic Camera}
Generally speaking, there are two types of bio-inspired
visual sampling manner: temporal contrast sampling and
integral sampling.
Dynamic Vision Sensor (DVS), also known as an event camera, generates events when pixel brightness changes exceed a threshold. 
Recently, Ev-NeRF \cite{enerf1} and EventNeRF \cite{enerf2} proposed neural radiance fields derived exclusively from event streams. However, due to the absence of texture details in event data, these approaches yield limited results. Ev-NeRF \cite{enerf1} is confined to learning grayscale NeRF, while EventNeRF \cite{enerf2} exhibits noticeable artifacts and chromatic aberration, lacking RGB data for supervision. 
\cite{qi2023e2nerf} first introduces event data and blurry RGB frames to achieve high quality clear scene reconstruction.
\cite{low2023robust} proposes Robust e-NeRF to robustly reconstruct NeRFs from moving event cameras under various real-world conditions,

Our work is based on spike camera, where each pixel responds independently to the accumulation of photons by generating spikes. It records full visual details with an ultra-high temporal resolution (up to 40kHZ). 
Image reconstruction based on spike cameras has emerged by analyzing spike intervals and counts \cite{recon}. Leveraging the spiking neuron model, a fovea-like texture reconstruction framework reconstructs images \cite{zhu2020retina}. Spike camera methods have been developed for tone mapping \cite{9181055}, motion deblurring \cite{Han_2020_CVPR}, and super resolution \cite{zhao2021super,xiang2021learning}. Recent deep learning-based image reconstruction methods \cite{zhao2021spk2imgnet,zhu2021neuspike,SSML,zhu2023recurrent} for spike cameras outperform traditional methods significantly. The noise characteristics of spike cameras differ substantially from those of traditional cameras. A preliminary analysis in \cite{zhu2021neuspike} demonstrates that the noise distribution varies under different lighting conditions. 

However, there is no established spike-based method for addressing the challenge of generating dense, photorealistic 3D representations of scenes.
The challenge is to explore the texture details hidden in the spike stream, understanding the spike distribution is crucial for spike-based applications.

\section{Methods}
\subsection{Neural Radiance Fields}
The fundamental idea of NeRF is to employ a Multilayer Perceptron (MLP) to learn a 3D volume representation~\cite{NeRF}. The MLP takes 3D position $\mathbf{o}$ and 2D ray direction $\mathbf{d}$ as inputs and produces color $\mathbf{c}$ and density $\sigma$ as outputs. In equation \eqref{eq:1}, $F_{\mathbf{\theta}}$ represents the MLP network with $\mathbf{\theta}$ as its parameters:
\begin{equation}
(\mathbf{c},\sigma) = F_{\mathbf{\theta}}(\gamma_{o}(\mathbf{o}),\gamma_{d}(\mathbf{d})).
\label{eq:1}
\end{equation}

The functions $\gamma_{o}(\cdot)$ and $\gamma_{d}(\cdot)$, defined in Equation \eqref{eq:1}, map the input 5D coordinates to a higher-dimensional space, enabling the neural network to better capture color and geometry information in the scene. 
To generate views from the scene representation, NeRF employs a classical volume rendering method described in Eq. \ref{eq:3}. For a given ray $\mathbf{r}(t)=\mathbf{o}+l\mathbf{d}$ emitted from the camera, the expected color projected on the pixel is denoted as $C(\mathbf{r})$. The near and far bounds of the ray are represented by $l_{n}$ and $l_{f}$, respectively. NeRF divides the interval $[l_{n}, l_{f}]$ into $N$ discrete bins, where $\mathbf{c}{i}$ and $\sigma{i}$ are the outputs of $F_{\mathbf{\theta}}$, indicating the color and density of each bin through which the ray passes. The distance between adjacent samples is $\delta_{i}=l_{i+1}-l_{i}$, and $T_{i}$ represents the transparency of particles between $l_{n}$ and bin $i$.
\begin{equation}
    \begin{aligned}
        C(\mathbf{r}) = \sum_{i=1}^{N}T_{i}(1-\exp(-{\sigma}_i{\delta}_i))\mathbf{c}_i,
    \end{aligned}
\label{eq:3}
\end{equation}
where $T(i) = \exp(-\sum_{j=1}^{i-1}\sigma_{j}\delta_{j})$.
To ensure reasonable sampling in the final model, NeRF utilizes a Hierarchical volume sampling strategy, simultaneously optimizing the coarse and fine models and using the density obtained by the coarse model to determine the sampling weight of the fine model.

\begin{figure*}[t]
  \centering
   \includegraphics[width=1\linewidth]{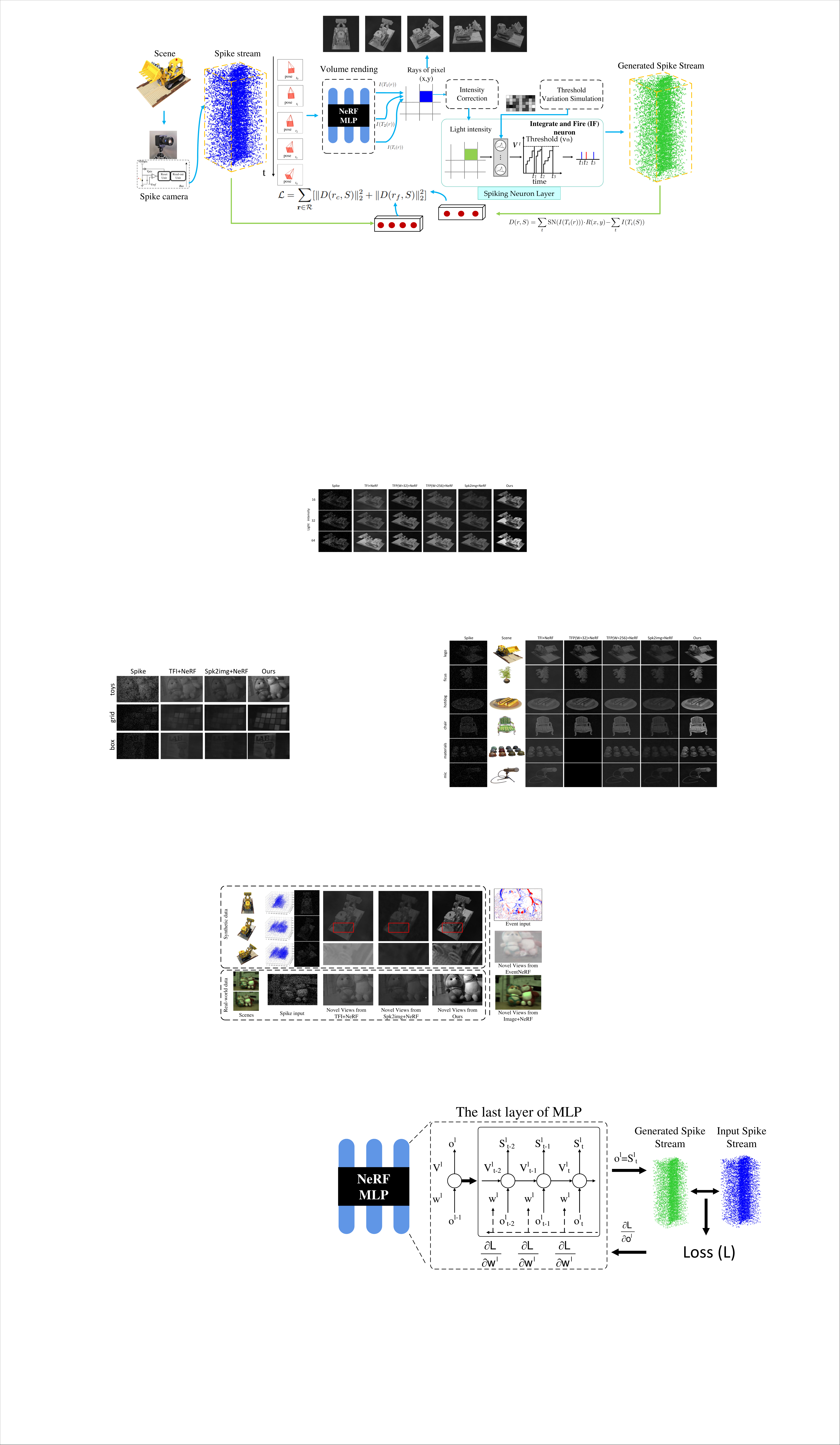}
   \caption{The architecture of SpikeNeRF. Motivated by the objective of learning NeRFs from a continuous spike stream, we establish the connection between the pixel ray $r$ and the real-world spike stream $S$. To quantify the rendering loss in the spike domain, we integrate a spiking neuron layer following the NeRF MLP. The nonuniformity is captured through pixel-to-pixel threshold variation, simulated by the spiking neuron layer. This tandem of loss functions ensures that the model can effectively capture and represent scene geometry. }
   \label{fig:framework}
\end{figure*}

\subsection{Spike Sampling Mechanism}
In the field of frame-free neuromorphic vision, the spike camera operates by converting the intensity of light into voltage through photoreceptors~\cite{recon,code}. Upon reaching a predefined threshold, a one-bit spike is generated, accompanied by a signal to reset the integrator. This mechanism resembles that of an integrate-and-fire neuron. Luminance stimuli $I$ lead to varying spike firing rates, with asynchronous triggering of output and reset across pixels. Brighter light results in faster firing, governed by the inequality $\int Idt \geq \phi$.

The raw data from the spike camera is a three-dimensional spike stream, focusing solely on luminance intensity integration and firing ultra-high frequency spikes. At each sampling moment, a digital signal of ``1'' (spike) or ``0'' is output based on whether a spike was fired. The spike firing status of pixel $(i,j)$ at moment $t$ is represented as $S_{i,j}(t) \in {0,1}$. 
The spike signal model is defined as:
\begin{equation}
S(x,y,t)=S(x,y,nt_0)=\epsilon\left(\int_{t'}^{n t_0}\frac{C,\Delta V}{I_{ph}(x,y,\tau)} d\tau-1\right),
\end{equation}
where $t'$ is the start time of the current integration phase, reset at the phase's end. $\Delta V$ denotes the voltage, $n t_0$ denotes the $n$-th clock signal, with $t_0$ as the unit time interval ($50\mu s$), and $I_{ph}$ represents the magnitude of the current for photoelectric conversion.
$C$ is the capacitor in parallel with the photodiode, which is used to store electrons for photo-conversion.

The noise of the spike camera differs significantly from traditional cameras due to circuit variations. The noise components encompass \emph{Shot Noise}, wherein the photosensor converts photons into photo-electrons, resulting in a photo-current subject to shot noise $N_{p}$ even in uniform light scenarios. \emph{Dark Current Noise} arises from thermal diffusion and defects, causing the spike circuit to produce signals even in the absence of scene light, with dark current noise denoted as $N_d$. \emph{Response Nonuniformity} stems from chip manufacturing limitations, causing pixel units to exhibit varying sensitivity to light intensity, leading to photo response nonuniformity noise $N_{nru}$. Lastly, \emph{Quantization Noise} arises from the delay in the release time of the spike signal relative to the generation time of the reset signal, introducing quantization noise $N_q$.

\subsection{Self-Supervision with Volumetric Rendering}
An overview of the training pipeline is illustrated in
Fig.~\ref{fig:framework}. In line with the spike generation model, we propose self-supervision with volumetric rendering in this section.
Different from traditional RGB frames, a binary spike frame can not accurately represent the light intensity at the current timestamp. 
Let us denote the intensity values of the pixel ray $r$ at time $T_i$ as $\hat{I}(T_i(r))$.
Since the high temporal resolution of the spike camera, the estimated short-term light intensity $\hat{L}_{T_i}$ can record motion details hidden in the spike stream. 
A straightforward approach is to calculate the light intensity $\hat{L}_{T_i}$ at time $T_i$ based on the original spike stream:
\begin{equation}
\hat{L}_{T_i} = \frac{\theta}{n-m}, \ \ m < n,
\end{equation}
where $m$ and $n$ are the timestamps corresponding to two adjacent spikes, $\theta$ controls the gray-scale level.

However, $\hat{L}_{T_i}$ is influenced by various sources of noise, particularly the temporal sampling noise \cite{zhu2021neuspike}. The firing intervals of spikes fluctuate even in constant illumination conditions due to the inherent spike sampling mechanism. For instance, in the integrate-and-fire mechanism, if the membrane potential of the neuron (pixel) is not reset, an object with a light intensity corresponding to an inter-spike interval of 2 might be recorded as 1 or 3.   
Consequently, $\hat{L}_{T_i}$ is suitable for capturing motion details in the intensity values of the pixel ray $r$, but texture details are often affected. $\hat{L}_{T_i}$ is susceptible to non-uniform pixel response, akin to standard image sensors. This is manifested by pixel-to-pixel variations in the contrast threshold, analogous to the spike camera's version of Fixed-Pattern Noise (FPN).
Hence, relying solely on the rendered and degraded estimated light intensity values for NeRF training supervision can yield suboptimal results.

\begin{figure}[t]
  \centering
   \includegraphics[width=\linewidth]{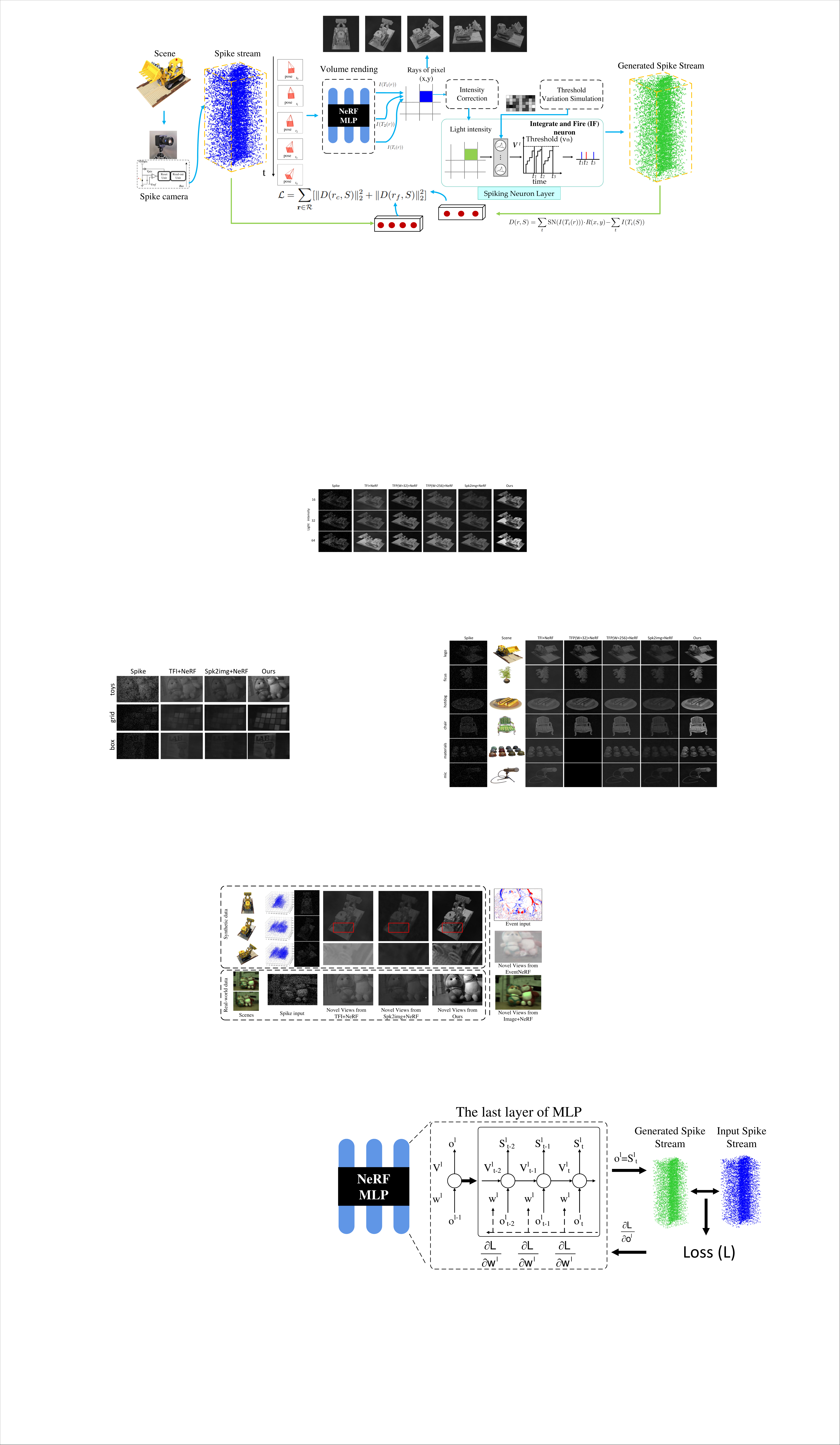}
   \vspace{-4mm}
   \caption{The backpropagation process of spiking neurons. The spiking neuron layer, comprising 256 time steps, follows the NeRF MLP. The weight of the last layer of MLP can be updated through Backpropagation Through Time (BPTT) using Eq. \ref{eq:neuron}.}
   \vspace{-4mm}
   \label{fig:back}
\end{figure}

\subsubsection{Measuring Rendering Loss in Spike Domain}
In order to capture the precise distribution of the spike stream, we advocate for employing a long-term spike rate rendering loss rather than directly assessing short-term light intensity. To account for the various noise sources, inspired by the spike simulator~\cite{zhu2023recurrent}, we introduce a layer of spiking neurons that generate spike streams based on the intensity values of the pixel ray $r$ emulating the spike sampling process. We then compute the long-term spike rate rendering loss directly on both the generated spike stream and the original spike stream.

Our findings indicate that by combining NeRF's inherent multi-view consistency with our long-term spike rendering loss, our framework can establish robust self-supervision. This approach proves effective in mitigating the impact of erroneous measurements, particularly in the presence of high noise levels and diverse illumination conditions.


\noindent\textbf{Relationship between the Pixel Ray $r$ and Real-world Spike Stream $S$.}
The objective is to estimate intensity values $\hat{I}(T_i(r))$ corresponding to the real light of the scene. 
Assuming that $\hat{I}(T_i(r))$ is clear, to supervise the $\hat{I}(T_i(r))$ with the real noisy spike stream, we need to consider multiple noises.
Inspired by~\cite{zhu2023recurrent}, we can define the relationship between $\hat{I}(T_i(r))$ and spike noise as:
\begin{equation}
    \hat{I}(T_i(r)) = {I}(T_i(S)) - {I}(T_i(N)),
\end{equation}
where ${I}(T_i(S)) = \frac{1}{\frac{Q_r}{L+N_{p}+N_{d}}+N_{rnu}+N_{q}}+N_{c}$, ${I}(T_i(N))$ denotes the intensity changes caused by noise, $L$ represents the scene light intensity, $Q_r$ is relative quantity matrix of electric charge.
$N_p$, $N_d$, $N_{rnu}$, $N_q$, and $N_c$ represent shot noise, dark current noise, response nonuniformity noise, quantization noise, and truncation noise, respectively.




\noindent\textbf{Spiking Neuron Layer.}
To accurately represent the spike distribution, we introduce an additional layer of spiking neurons following the NeRF MLP to generate the spike stream. Given that the spike camera employs an integrate-and-fire mechanism to convert the light intensity of the scene into spike streams, our framework utilizes the Integrate-and-Fire (IF) neuron model. The Integrate-and-Fire (IF) model~\cite{gerstner2014neuronal} is a widely employed neuron model in Spiking Neural Networks (SNN) known for its biological realism~\cite{wu2021progressive}.
\begin{figure*}[t]
  \centering
   \includegraphics[width=1\linewidth]{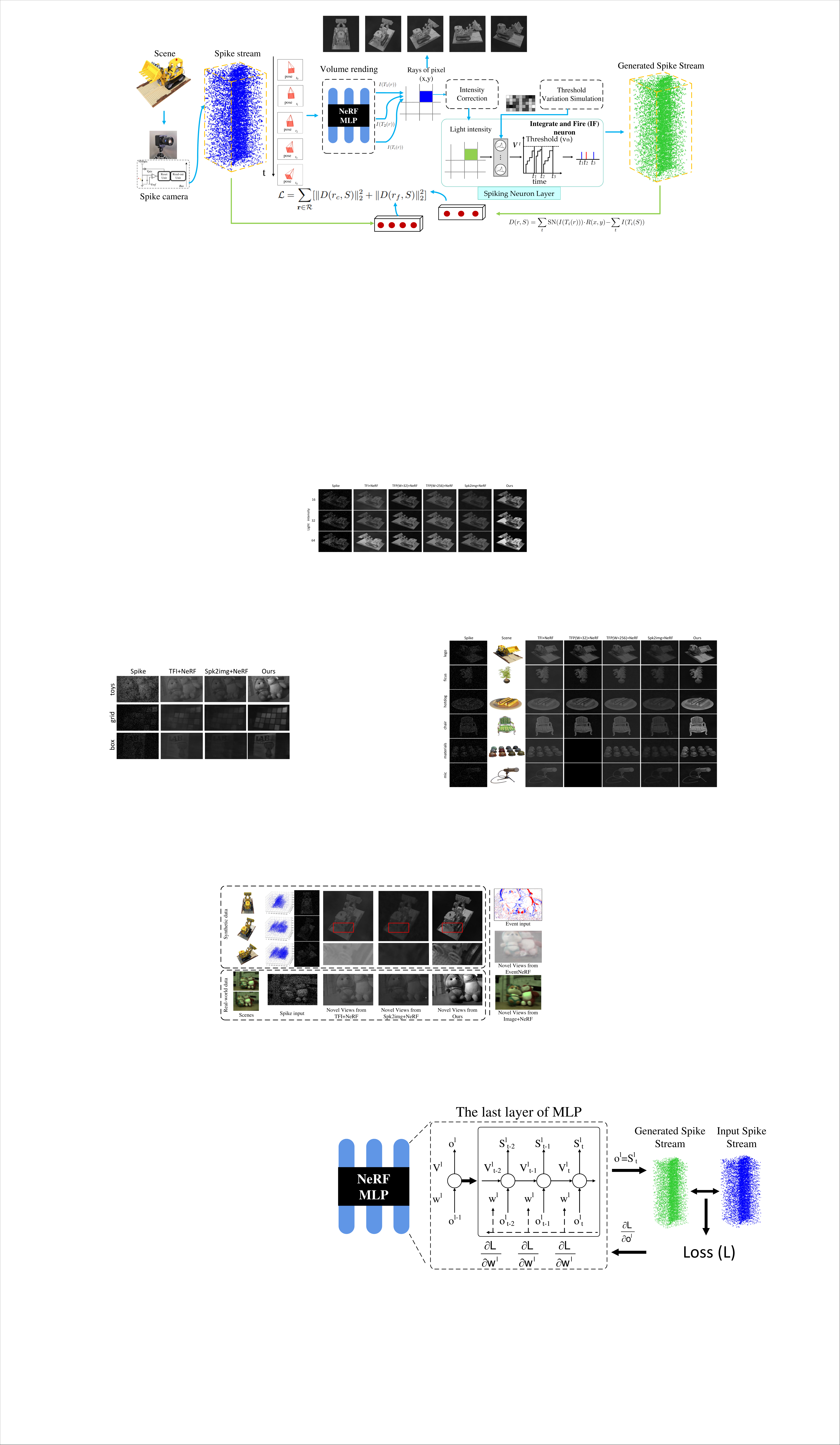}

   \caption{Quantitative results on synthetic spike data.}
   \label{fig:exp1}
\end{figure*}
For better representation, we represent the function of IF neuron as the discrete form:
\begin{eqnarray}
V_t = V_{t-1} + X_t, \ \ 
S_t = H(V_t - V_{th}),
\label{eq2}
\end{eqnarray}
where $V(t)$ denotes the membrane potential after neuronal dynamics at $t$, $X(t)$ represents the input to neuron. A spike fires if $V(t)$ exceeds the threshold $V_{th}$, $V_{rest}$ is the resting potential after firing, $S_t$ denotes the spike output at $t$, $H(\cdot)$ denotes the Heaviside step function which is defined as $H(x) = 1$ for $x \geq 0$ and $H(x) = 0$ for $x < 0$.

\noindent\textbf{Threshold Variation Simulation.}
The response nonuniformity noise can be characterized by the pixel-to-pixel variation in the firing threshold.
In fact, the deviation matrix corresponding to the response nonuniformity noise can be obtained by capturing a uniform light scene and recording the intensity.
Thus, the threshold variation of each spiking neuron can be modeled by Eq.~\ref{eq:th}, and the spike stream can be generated by
\begin{equation}
    \hat{S}(x,y) = {\rm SN}({I}(T_i(r))) \cdot R(x,y),
    \label{eq:th}
\end{equation}
where ${\rm SN}(\cdot)$ denotes the spiking neuron, $R(x,y)$ is the nonuniformity matrix and can be obtained by capturing a uniform light scene and recording the intensity\footnote{Choose the pixel ($x_m$, $y_m$) which is closest to the average response value as the reference pixel. $R(x,y)$ is then obtained by calculating the ratio of the reference pixel's response value to the response values of other pixels: $R(x,y)=\frac{(L_2+L_d(x_m, y_m))T_2(x_m,y_m)}{(L_2+L_d(x, y))T_2}$, where $L$ and $T$ are variables to be calibrated. See our supplementary materials for details.}.

\noindent\textbf{Long-term Spike Rendering Loss.}
The spike stream can be mapped into an RKHS (reproducing kernel Hilbert space) as continuous-time functions which incorporate a statistical description of spike trains.
Based on the RKHS, the kernel method~\cite{dong2018spike} is employed for measuring spike train distances. Additionally, \cite{zhu2020hybrid} introduces an intensity-based distance metric to quantify spike distances. Nevertheless, the complexity of these two methods is excessively high, rendering NeRF training inefficient.

In order to measure the distance between the generated spike stream $\hat{S}$ and the input spike stream $S$, we propose a long-term spike rendering loss, which can be expressed by 
\begin{equation}
     D(r,S) = \sum_t{\rm SN}({I}(T_i(r))) \cdot R(x,y)-\sum_t{I}(T_i(S)).
\label{eq:8}
\end{equation}

In theory, employing a longer spike stream for measurement has the potential to mitigate excessive truncation noise. In our experiment, the parameter $t$ is configured to 256, aligning with the time steps of the spiking neuron. The final loss function can be expressed as follows:

\begin{equation}
    \mathcal{L}= \sum_{\mathbf{r}\in\mathcal{R}}[\|D(r_c,S) \|_{2}^{2}+\\
    \|D(r_f,S)\|_{2}^{2}],
\label{eq:8}
\end{equation}
where $r_c$ and $r_f$ are the pixel rays of coarse and fine models, respectively.
In this way, the loss function of NeRF with spike stream as input is converted into Eq.~\ref{eq:8}. We adopt the design of the joint optimization of NeRF's coarse model and fine model, which is still beneficial in our framework.

\subsubsection{Backpropagation Training}
The spike generation function of an IF neuron is a hard threshold function that emits a spike when the membrane potential exceeds a firing threshold. Due to this discontinuous and non-differentiable neuron model, standard backpropagation algorithms cannot be applied to SNNs in their native form~\cite{zhang2018highly,zhang2021rectified}. 
As shown in Fig.~\ref{fig:back}, the backpropagated errors pass through the spiking neuron layer using BackPropagation Through Time (BPTT)~\cite{werbos1990backpropagation}. 
Since the spiking neurons are one-to-one connected to the neurons of the last layer of NeRF MLP, the weight of the last layer of NeRF MLP $w^{{l}}$ is updated by BPTT.
In BPTT, the network is unrolled for all discrete time steps, and $w^{{l}}$ update is computed as the sum of gradients from each time step as follows:
{\small
\begin{eqnarray}
\Delta w ^{{l}} = \sum_{t-1}^{T} \frac{\partial \mathcal{L}_{total}}{\partial o^l_t}  \frac{\partial o^l_t}{\partial V^l_t} \frac{\partial{V}^l_t}{\partial w^{{l}} },  
{\rm where}\frac{\partial o^l_t}{\partial V^l_t}=H_1^{'}(V_t-V_{th}),
\label{eq:neuron}
\end{eqnarray}
}
where $o^l_t$ is the output of the neuron at time $t$, $\frac{\partial o^l_t}{\partial V^l_t}$ denotes the derivative of spike with respect to the membrane potential after charging at time step $t$. Since $\frac{\partial o^l_t}{\partial V^l_t}$ is not differentiable, we adopt surrogate gradient method~\cite{neftci2019surrogate} to calculate it.
In our work, the approximate IF gradient is computed as $H_1(x) =\frac{1}{V_{th}}$, where the threshold value accounts for the change of the spiking output with respect to the input.

\begin{table}[t]
\small
\centering
\caption{Quantitative evaluation on synthetic dataset.}
\label{table1}
\begin{tabular}{cccc}
\toprule[1pt]
Method / Metric & PSNR $\uparrow$ & SSIM $\uparrow$ & LPIPS $\downarrow$\\
\hline
TFP (32) + NeRF&   16.21/16.63&0.142/0.648&0.662/0.058 \\
TFP (256) + NeRF& 15.07/19.24&0.094/0.864& 0.741/0.085  \\ 
TFI + NeRF&  14.93/19.34 & 0.108/0.862 & 0.752/0.090 \\
Spk2img + NeRF&  13.41/14.05 & 0.064/0.749& 0.729/0.133\\
Ours& \textbf{20.78}/\textbf{22.07} & \textbf{0.209}/\textbf{0.916} & \textbf{0.617}/\textbf{0.053}\\ \toprule[1pt]
\end{tabular}
\label{table1}
\end{table}

Based on the above, a learnable realistic noise-embedded spike stream generation pipeline is proposed.
Each time step of the spiking neuron represents a readout circuit process that verifies the reset signal in a clock signal corresponding to a spike camera. 
The dark current noise, nonuniformity noise, and quantization noise are collectively treated as nonuniformity noise. These components are incorporated into the generated spike stream through a simulation of threshold variation.

\section{Experiment}

\begin{figure}[t]
  \centering
   \includegraphics[width=1\linewidth]{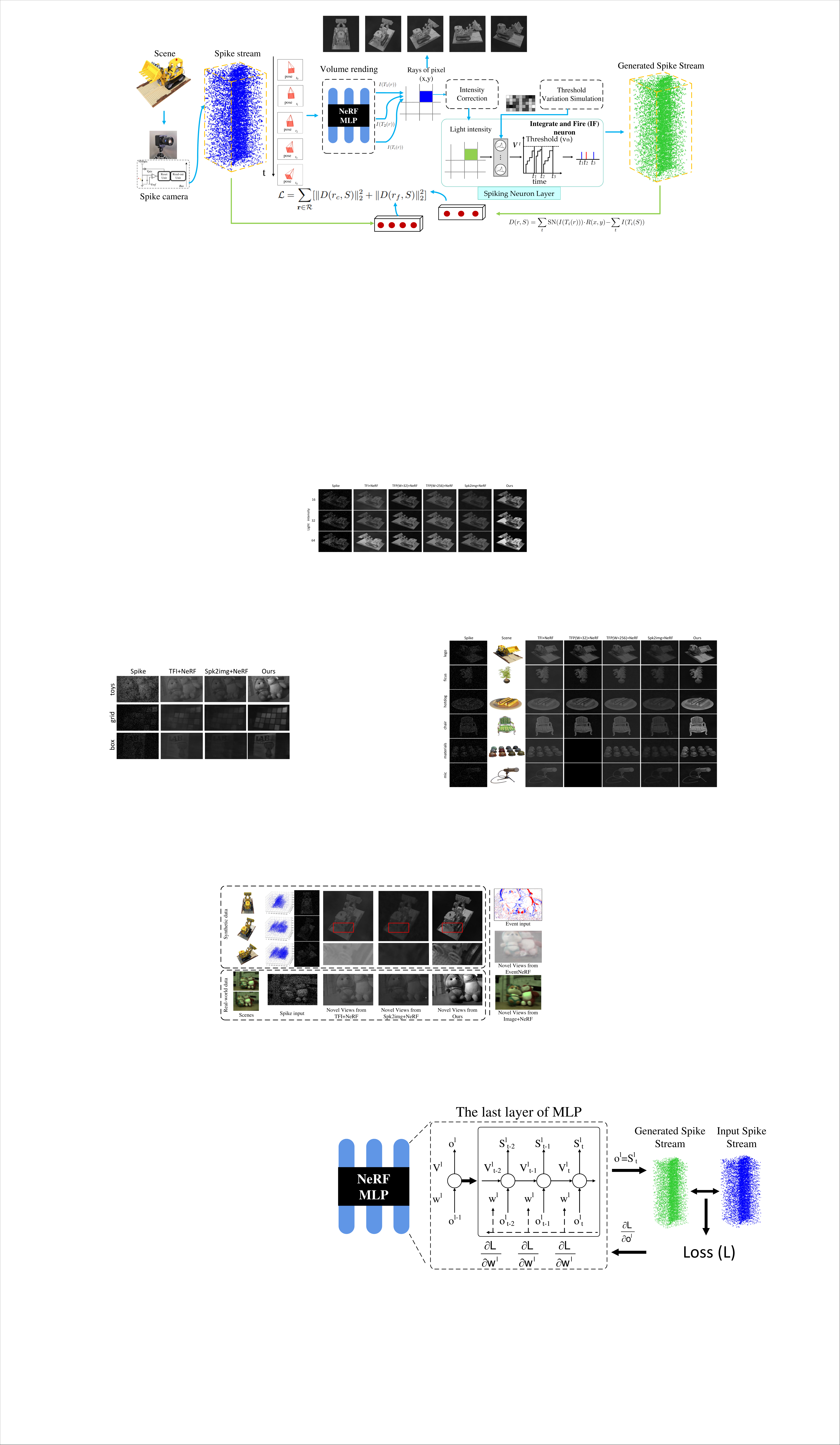}

   \caption{Quantitative results on real-world spike data.}
   \label{fig:real}
\end{figure}
\begin{figure*}[t]
  \centering
   \includegraphics[width=\linewidth]{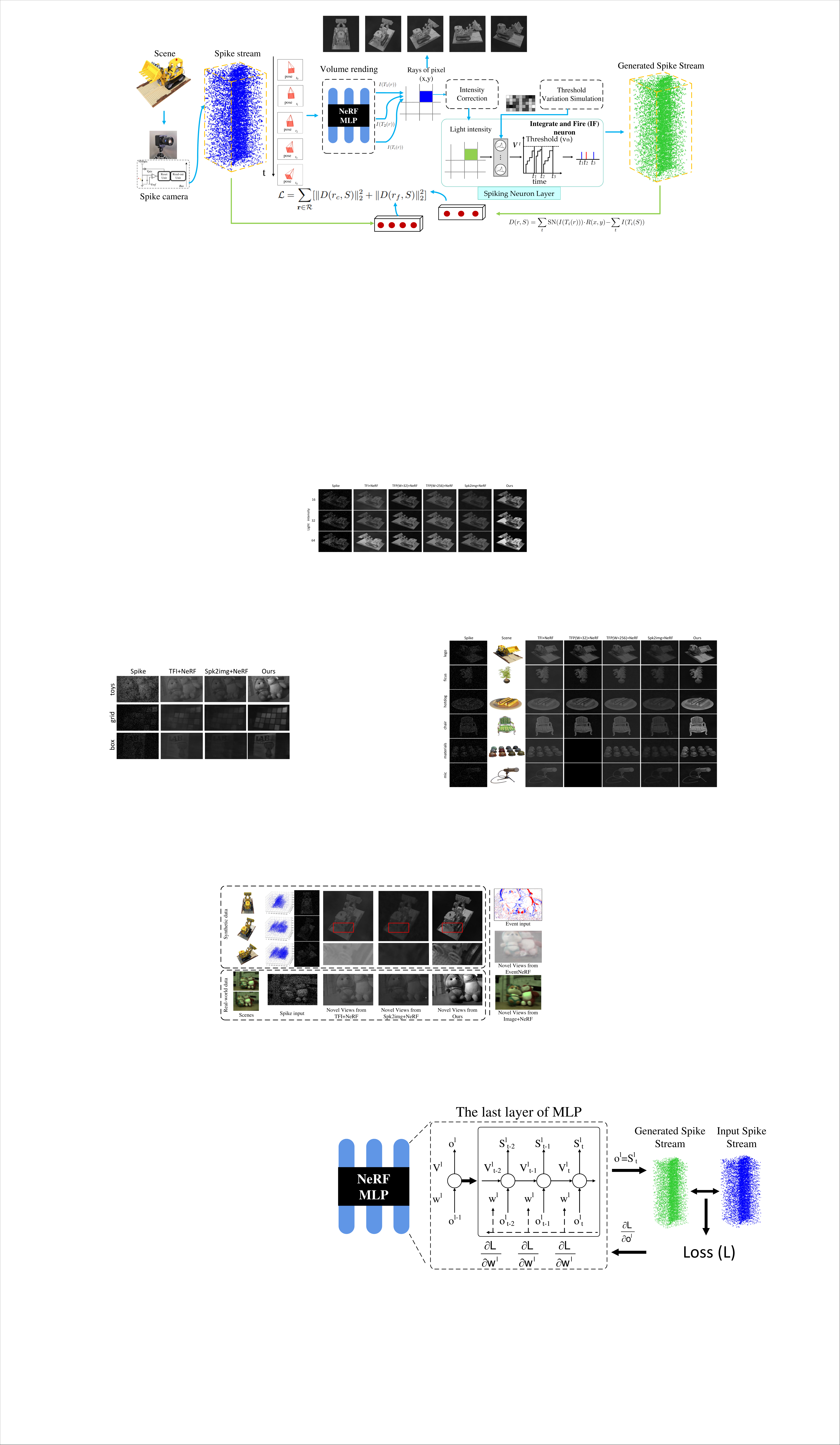}
   \vspace{-4mm}
   \caption{Quantitative results on different light intensities.}
   \label{fig:intensity}
\end{figure*}

\begin{table*}[t]
\tabcolsep=1.2mm
\footnotesize
\centering
\caption{Ablation study on synthetic dataset.}
\vspace{-4mm}
\begin{tabular}{cccccccccccccc}
\toprule[1pt]
\multirow{2}{*}{Method}&\multirow{2}{*}{Loss}&\multicolumn{3}{c}{Static} &\multicolumn{3}{c}{Dynamic} &\multicolumn{3}{c}{Average}\\
\cmidrule(r){3-5} \cmidrule(r){6-8} \cmidrule(r){9-11}

&  & PSNR $\uparrow$ & SSIM $\uparrow$ & LPIPS $\downarrow$& PSNR $\uparrow$ & SSIM $\uparrow$ & LPIPS $\downarrow$& PSNR $\uparrow$ & SSIM $\uparrow$ & LPIPS $\downarrow$\\
\hline
TFP(32)+NeRF& MSE&   15.26/15.57 &0.154/0.785 & 0.588/0.065 & 14.93/15.22& 0.112/0.745&0.626/0.109&15.10/15.40&0.133/0.765& 0.607/0.087 \\
TFP(256)+NeRF& MSE& 14.44/16.41 & 0.127/0.794 & 0.652/0.088 & 14.34/16.29 & 0.099/0.770& 0.693/0.114& 14.39/16.35 & 0.113/0.782& 0.672/0.101\\ 
TFI+NeRF& MSE& 13.77/16.63& 0.120/0.793&0.670/0.093& 13.76/16.11& 0.096/0.769& 0.694/0.110 & 13.76/16.37& 0.108/0.781 & 0.682/0.101\\
Spk2img+NeRF& MSE& 13.42/14.05& 0.066/0.726& 0.724/0.129& 13.39/14.04& 0.062/0.772& 0.735/0.137& 13.41/14.05& 0.064/0.075& 0.729/0.133\\ \hline
\multirow{4}{*}{Ours}& $\mathcal{L}_i$$^{*}$& 13.77/16.63& 0.120/0.793&0.670/0.093& 13.76/16.11& 0.096/0.769& 0.694/0.110 & 13.76/16.37& 0.108/0.781 & 0.682/0.101\\
& $\mathcal{L}_i$& 18.55/\textbf{19.76}& 0.237/0.878& 0.527/0.051& 16.63/17.79& 0.167/0.797& 0.593/0.097& 17.59/18.77& 0.202/0.837& 0.560/0.074 \\ 

&$\mathcal{L}_s$+$\mathcal{L}_i$& 18.50/19.59& 0.237/0.876& 0.524/0.051& 17.61/18.41& 0.179/0.816& \textbf{0.590}/\textbf{0.089}& 18.05/19.00& 0.208/0.846& \textbf{0.557}/\textbf{0.070}\\ 
&$\mathcal{L}_s$& \textbf{18.72}/19.38& \textbf{0.251}/\textbf{0.880}& \textbf{0.517}/\textbf{0.050}& \textbf{18.16}/\textbf{18.80}& \textbf{0.187}/\textbf{0.820}& 0.601/0.093& \textbf{18.44}/\textbf{19.09}& \textbf{0.219}/\textbf{0.850}& 0.559/0.072\\\toprule[1pt]
\end{tabular}
\label{table2}
\end{table*}

\begin{table*}[t]
\footnotesize
\tabcolsep=1.72mm
\centering
\caption{Quantitative evaluation of different light intensities on synthetic dataset.}
\vspace{-4mm}
\begin{tabular}{ccccccccccccc}
\toprule[1pt]
\multirow{2}{*}{Method}&\multicolumn{3}{c}{Light intensity (16)} &\multicolumn{3}{c}{Light intensity (32)} &\multicolumn{3}{c}{Light intensity (64)}\\
\cmidrule(r){2-4} \cmidrule(r){5-7} \cmidrule(r){8-10}

& PSNR $\uparrow$ & SSIM $\uparrow$ & LPIPS $\downarrow$& PSNR $\uparrow$ & SSIM $\uparrow$ & LPIPS $\downarrow$& PSNR $\uparrow$ & SSIM $\uparrow$ & LPIPS $\downarrow$\\
\hline
TFP(32)+NeRF& 15.27/15.57& 0.154/0.785& 0.588/0.065& 17.81/18.55& 0.191/0.827& 0.562/0.077& 20.63/21.86& 0.239/0.872& 0.528/0.067  \\
TFP(256)+NeRF& 14.44/16.41&0.127/0.794& 0.652/0.088& 15.82/18.68& 0.170/0.835& 0.616/0.077& 17.28/21.27& 0.189/0.852& 0.596/0.081  \\ 
TFI+NeRF& 13.77/16.63& 0.120/0.793& 0.670/0.093& 15.07/18.53& 0.156/0.825& 0.634/0.083& 16.16/20.56& 0.187/0.852& 0.602/0.077\\
Spk2img+NeRF& 13.42/14.05& 0.066/0.726& 0.724/0.129& 14.59/15.39& 0.109/0.768& 0.651/0.093& 16.37/17.73& 0.169/0.837& 0.595/0.073\\ \hline
Ours& \textbf{18.72}/\textbf{19.38}& \textbf{0.251}/\textbf{0.880}& \textbf{0.517}/\textbf{0.050}& \textbf{22.05}/\textbf{23.66}& \textbf{0.300}/\textbf{0.926}& \textbf{0.477}/\textbf{0.050}& \textbf{23.89}/\textbf{24.46}& \textbf{0.411}/\textbf{0.929}& \textbf{0.428}/\textbf{0.049}\\ \toprule[1pt]
\end{tabular}
\label{table3}
\end{table*}

\subsection{Experimental Setup}
\noindent\textbf{Synthetic Spike Data.} 
We utilize six synthetic scenes (chair, ficus, hotdog, lego, materials, and mic) to generate synthetic spike data in NeRF. Using the spike generator provided by \cite{zhu2023recurrent}, we simulate spike streams for each viewpoint under different illumination conditions. The original image size is resized to 400 $\times$ 400, resulting in generated spike streams with a resolution of $400\times400\times256$, where 256 represents the time length of each spike stream.
In addition to generating spike streams from static viewpoint images, we render 16 high-resolution images captured by the camera and feed them into the spike generator to replicate the dynamic recording process of the spike camera. Each scene comprises 100 sets of images and their corresponding event data. Further details about the synthetic data can be found in our supplementary material.

 

\noindent\textbf{Real-world Spike Data.} 
We employ the spike camera~\cite{huang20231000} to record real-world spike data. This camera is capable of capturing spike streams with a spatial resolution of $250\times400$ and a temporal resolution of 20,000 Hz. Holding the camera by hand, we capture data from five real-world scenarios, each featuring texture details under different illumination conditions. Each dataset comprises approximately 35 images from various viewpoints along with their corresponding spike data.
To facilitate a comprehensive comparison of different cameras, we also use the DAVIS346 color event camera~\cite{davis346} to capture event and RGB data. This camera is equipped to capture spatial-temporally aligned event data and RGB frames, with a resolution of 346$\times$260 and an exposure time set to 33ms for the RGB frames.

\noindent\textbf{Baselines.}
Given the absence of existing spike-based NeRF methods, we compare our approach with three spike-based image reconstruction methods: TFI~\cite{recon,zhu2022ultra}, TFP~\cite{recon,zhu2022ultra}, and Spk2imgNet~\cite{zhao2021spk2imgnet}. In a two-stage process, we reconstruct images using TFI and TFP  (with window sizes of 32 and 256), which are commonly employed spike-based reconstruction methods, and Spk2imgNet, a learning-based method leveraging neural networks.
While there are other learning-based methods for spike-based reconstruction~\cite{SSML}, those methods aim to simultaneously estimate flow and image. The approach by \cite{zhu2023recurrent} requires long spike sequences for recurrently optimizing the reconstruction process. Consequently, Spk2imgNet is well-suited for the NeRF setting, demonstrating stable performance.
To ensure a fair comparison, all methods adopt a common NeRF backbone.


\noindent\textbf{Training Details.}
For the simulated data, we use the poses provided by the Blender and the generated spike stream as the input of the NeRF framework.
For real-world data, we first reconstruct the spike stream into images, and then use COLMAP to estimate the poses.  For a fair comparison, all comparison methods use the same poses as the input of the network.
At each optimization iteration, we randomly sample a batch of camera rays from the set of all pixels in the dataset. 
We use the Adam optimizer with a learning rate that  begins at 5 $\times$ 
$10^{-4}$ and decays exponentially to 5 $\times$ 10$^{-5}$.
The optimization for a single scene typically takes about 20 hours to converge on a single NVIDIA 3090 GPU.

\begin{table}[t]
\small
\centering
\caption{Quantitative evaluation on real-world data.}
\begin{tabular}{cccccc}
\toprule[1pt]
Camera & Method & Brisque $\downarrow$ & NIQE $\downarrow$\\
\hline
APS & Image+NeRF& 52.07& 9.06 \\
APS & Image+NeRF$^*$&  53.05& 10.06\\
DVS & Event+ENeRF&  83.98& 12.71\\ 
Spike camera& TFP+NeRF& 42.64& 6.16\\
Spike camera& TFI+NeRF& 47.92& 7.49\\
Spike camera& Ours & \textbf{35.94}& \textbf{6.09} \\ \toprule[1pt]
\end{tabular}
\label{table4}
\end{table}

\subsection{Quantitative Experiment}
We utilize three standard metrics: PSNR, SSIM~\cite{wang2004image}, and LPIPS~\cite{zhang2018unreasonable}. These metrics are employed to quantify the similarity between the synthesized novel views and the provided target novel views.
The quantitative results on the synthetic dataset are presented in Table \ref{table1}. Since only the object will produce firing spikes in the synthetic dataset, we provide two measurement approaches: the left one utilizes the entire image for measurement, while the right one focuses solely on measurements within the object region.
The results show that our model performs better than other two-stage methods, proving the effectiveness of the proposed framework.
Table \ref{table4} presents the results on real-world data. As the real data lacks ground truth, we assess the results using no-reference quality assessment methods, specifically Brique~\cite{mittal2012no} and NIQE~\cite{mittal2012making}. The results indicate that our method outperforms other models. Additional experimental results can be found in our supplementary materials.

\subsection{Qualitative Experiment}
We perform qualitative experiments on both synthetic and real-world data. As shown in Fig. \ref{fig:intensity}, the rendered views from our method show markedly improved sharpness.
TFP (W=32) struggles to generate novel views for the ``mic" and ``materials" scenes due to their low light intensity. The short window used by TFP fails to gather sufficient information.
Similarly, Spk2img also exhibits suboptimal performance. This learning-based method is designed for ideal illumination conditions, and its performance sharply degrades when confronted with noisy spikes.
In Fig. \ref{fig:real}, three real-world scenes with varying illumination conditions are depicted. ``toys" exhibits high light intensity, ``grid" features medium light, and ``box" has low illumination. The results illustrate that our model performs effectively across different scenes and lighting conditions in real spike data.

\subsection{Ablation Study}
We perform ablation studies on the proposed long-term spike rendering loss, and the results are presented in Table \ref{table2}. In the table, $\mathcal{L}_s$ represents the proposed spike rendering loss, $\mathcal{L}_i$ involves using the light intensity of rays plus threshold correction, and the reconstructed images from spikes as supervision, while $\mathcal{L}_i^*$ directly employs the estimated light intensity and reconstructed images as supervision.
The experiments are conducted in both static and dynamic settings, affirming the effectiveness of our framework.
Furthermore, we assess the performance under varying light intensities on the synthetic dataset. The results, presented in Table \ref{table3} and Fig. \ref{fig:intensity}, demonstrate the robustness of our framework across different illumination conditions.

\section{Conclusion}
This paper introduces SpikeNeRF as a pioneering approach to derive a volumetric scene representation from spike camera data using NeRF. With a focus on purely spike-based supervision, SpikeNeRF preserves texture and motion details in high temporal resolution, addressing the challenges associated with real-world spike sequences. Our evaluations on a newly curated dataset of synthetic and real spike sequences demonstrate the efficacy of SpikeNeRF for novel view synthesis. We hope that our work will shed light on the research of high quality 3D representation learning with novel spike stream techniques.

\noindent\textbf{Acknowledgement}
This work was supported by National Natural Science Foundation of China under Grant No.62302041, China National Postdoctoral Program under contract No. BX20230469, and Beijing Institute of Technology Research Fund Program for Young Scholars.

{
    \small
    \bibliographystyle{ieeenat_fullname}
    \bibliography{main}
}

\clearpage
\setcounter{page}{1}
\maketitlesupplementary
\renewcommand\thefigure{S\arabic{figure}}
\setcounter{figure}{0}
\renewcommand\thetable{S\arabic{table}}
\setcounter{table}{0}
\renewcommand\theequation{S\arabic{equation}}
\setcounter{equation}{0}

\section{Synthetic Data Details}
We generate synthetic spike data in NeRF using six scenes (chair, ficus, hotdog, lego, materials, and mic). To synthesize the spike stream, we initially resize the original images from Blender to 400 $\times$ 400. Employing the spike generator provided by \cite{zhu2023recurrent}, we simulate spike streams for each viewpoint.

For varying illumination conditions, we manipulate the intensity parameter in the simulator within the range of 16 to 64 (refer to Fig. 6 of the main manuscript and Table~\ref{tab:s1}). Beyond generating spike streams from static viewpoint images, we also render 16 high-resolution images captured by the camera. These images are then input into the spike generator to replicate the dynamic recording process of the spike camera. The resulting spike streams for each view are of size 400 $\times$ 400 $\times$ 256.

\begin{figure}[t]
  \centering
   \includegraphics[width=1\linewidth]{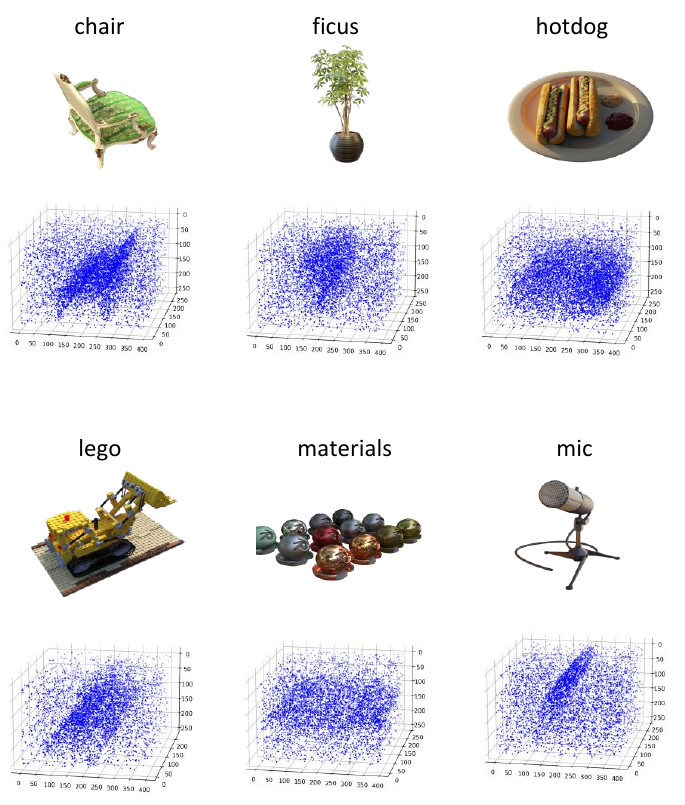}

   \caption{The input images and the visualization of the generated spike stream in synthetic scenes.}
   \label{fig:s1}
\end{figure}

Each scene encompasses 100 sets of images and their corresponding event data. The input images from Blender and the generated spike streams are visually depicted in Fig.~\ref{fig:s1}.



\section{Real-world Data Details}
The spike camera is capable of capturing spike streams with a spatial resolution of $250\times400$ and a temporal resolution of 20,000 Hz. 

\begin{figure*}[t]
  \centering
   \includegraphics[width=1\linewidth]{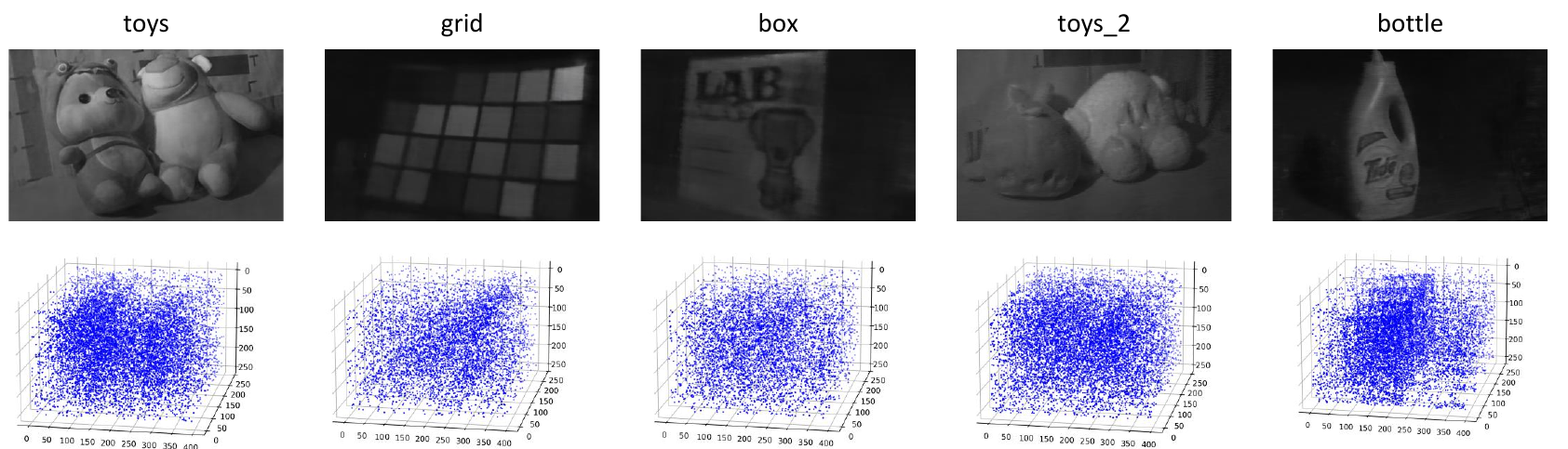}

   \caption{The scenes and the visualization of the recorded spike stream in real-world settings.}
   \label{fig:s2}
\end{figure*}

For each viewpoint, we simultaneously capture ideal and non-ideal conditions of spike data. Initially, we minimize noise by providing the spike camera with ideal light intensity and motion. Utilizing the Spk2img method, we reconstruct high-quality images and employ COLMAP for pose estimation. This process enables us to obtain the pose and camera parameters for spike data under typical conditions.
Conducting handheld captures, we gather data from five real-world scenarios, each showcasing texture details under distinct illumination conditions. Each dataset consists of approximately 35 images from diverse viewpoints, accompanied by their corresponding spike data.
The recorded spike streams are visually depicted in Fig.~\ref{fig:s2}. The spike numbers of both synthetic spike data and real-world spike data are shown in Table~\ref{tab:s1}.

\section{More Details of Threshold Variation Simulation}
The threshold variation of each spiking neuron can be modeled by Eq.~\ref{eq:th}, and the spike stream can be generated by
\begin{equation}
    \hat{S}(x,y) = {\rm SN}({I}(T_i(r))) \cdot R(x,y),
    \label{eq:th}
\end{equation}
where ${\rm SN}(\cdot)$ denotes the spiking neuron, $R(x,y)$ is the nonuniformity matrix and can be obtained by capturing a uniform light scene and recording the intensity. 

\begin{table*}[t]
\small
\centering
\caption{The mean spike count for each view across synthetic and real-world datasets.}
\begin{tabular}{ccccccc}
\toprule[1pt]
\multicolumn{7}{c}{Synthetic spike data} \\
\cmidrule(r){1-7} 
$\#$ SpikeNum/Scene & Lego & Chair&Hotdog & Ficus& Materials & Mic\\
\hline
Train&   268,185&	262,640&	282,359	&190,502	&196,675	&169,442 \\
Test& 264,969&	264,822	&255,235	&197,473&210,756	&167,507 \\ 
Val&  270,027&	257,689&	274,341	&192,482&	198,026	&169,398
\\ \toprule[1pt]

\multicolumn{7}{c}{Real-world spike data} \\
\cmidrule(r){1-7} 
$\#$ SpikeNum/Scene & Toy & Grid & Box & Toy$\_$2& Bottle&- \\
\hline
-&  585,016&	422,037&	441,255&	725,272&691,757	&-
\\ \toprule[1pt]

\end{tabular}
\label{tab:s1}
\end{table*}

\begin{table*}
  \centering
  \small
  \setlength{\tabcolsep}{0.8mm}
  \renewcommand{\arraystretch}{1.2}
    \caption{Detailed quantitative result on six synthetic scenes. We use \textbf{bold} to mark the best results. The results on the left pertain to measurements across the entire image, while those on the right specifically focus on measurements within the object region.}
  \label{tab:s2}
  \begin{tabular}{l|ccc|ccc|ccc}
    \toprule
    
    & \multicolumn{3}{c}{Lego} & \multicolumn{3}{c}{Chair}  & \multicolumn{3}{c}{Hotdog}\\

    {Novel View}& PSNR$\uparrow$ &SSIM$\uparrow$ &LPIPS$\downarrow$ & PSNR$\uparrow$ &SSIM$\uparrow$ &LPIPS$\downarrow$ & PSNR$\uparrow$ &SSIM$\uparrow$ &LPIPS$\downarrow$ \\
    
    \midrule
     TFP(32)+NeRF& 15.27/15.57& 0.154/0.785& 0.588/0.065&	14.94/15.21&	0.126/0.844	&0.662/0.057&	16.18/16.56	&0.191/0.853&	0.641/0.059\\
TFP(256)+NeRF& 14.44/16.41&0.127/0.794& 0.652/0.088 &13.92/15.47	&0.089/0.836&	0.735/0.079	&15.15/17.34	&0.175/0.870&	0.685/0.059 \\ 
TFI+NeRF& 13.77/16.63& 0.120/0.793& 0.670/0.093&	13.20/15.83	&0.085/0.835&	0.753/0.090	&14.48/17.56&	0.172/0.871&	0.700/0.063\\
Spk2img+NeRF& 13.42/14.05& 0.066/0.726& 0.724/0.129&	12.76/13.23&	0.055/0.794&	0.778/0.102&	14.20/14.92	&0.123/0.810	&0.728/0.087\\ \hline
Ours& \textbf{18.72}/\textbf{19.38}& \textbf{0.251}/\textbf{0.880}& \textbf{0.517}/\textbf{0.050}&	\textbf{19.11}/\textbf{19.76}&	\textbf{0.201}/\textbf{0.917}	&\textbf{0.591}/\textbf{0.045}&	\textbf{19.76}/\textbf{20.25}	&\textbf{0.274}/\textbf{0.930}	&\textbf{0.581}/\textbf{0.030} \\

    \toprule
    & \multicolumn{3}{c}{Ficus} & \multicolumn{3}{c}{Materials}  & \multicolumn{3}{c}{Mic}\\

    {Novel View}& PSNR$\uparrow$ &SSIM$\uparrow$ &LPIPS$\downarrow$ & PSNR$\uparrow$ &SSIM$\uparrow$ &LPIPS$\downarrow$ & PSNR$\uparrow$ &SSIM$\uparrow$ &LPIPS$\downarrow$ \\
    
    \midrule
     TFP(32)+NeRF& 18.44/19.17& 0.096/0.876& 0.758/0.050&	 / &	/ 	& / &	/ 	& / &	 / \\
TFP(256)+NeRF& 16.55/20.45&0.067/0.894& 0.814/0.049 &11.52/20.63	&0.026/0.855&	0.823/0.130	&17.85/25.14	&0.080/0.937&	0.817/0.098 \\ 
TFI+NeRF& 15.72/20.29& 0.061/0.891& 0.829/0.052&	15.92/20.70	&0.138/0.849&	0.726/0.139	&16.48/25.02&	0.070/0.934&	0.837/0.103\\
Spk2img+NeRF& 16.33/17.60& 0.037/0.848& 0.872/0.070&	16.62/17.78&	0.092/0.795&	0.782/0.166&	18.94/21.86	&0.049/0.901	&0.889/0.121\\ \hline
Ours& \textbf{20.89}/\textbf{22.18}& \textbf{0.132}/\textbf{0.910}& \textbf{0.725}/\textbf{0.044}&	\textbf{21.97}/\textbf{23.14}&	\textbf{0.244}/\textbf{0.903}	&\textbf{0.595}/\textbf{0.077}&	\textbf{24.21}/\textbf{27.43}	&\textbf{0.149}/\textbf{0.953}	&\textbf{0.693}/\textbf{0.053}
\\
    
    \bottomrule
  \end{tabular}

\end{table*}

Choosing the pixel ($x_m$, $y_m$) which is closest to the average response value as the reference pixel, $R(x,y)$ is then obtained by calculating the ratio of the reference pixel's response value to the response values of other pixels: $R(x,y)=\frac{(L_2+L_d(x_m, y_m))T_2(x_m,y_m)}{(L_2+L_d(x, y))T_2}$, where $L_2$ and $T_2$ are variables to be calibrated. 
As referred to \cite{zhu2023recurrent}, fixed pattern noise includes dark current noise and response nonuniformity noise, the equivalent light intensity value for the dark signal, $L_d$ can be calculated by capturing two uniformly illuminated scenes:
\begin{equation}
	\begin{cases}
	C\Delta V = \alpha L_d T_d\\
	C\Delta V = \alpha (L_1 + L_d)T_1
		   \end{cases},
    \label{eq:s2}
\end{equation}
where the first line of the equation represents capturing the scene brightness at zero (obtained by covering the lens in a dark room), while the second line represents capturing the scene brightness at $L_1$ (recorded using a photometer for $L_1$ value).
$T_d$ and $T_1$ respectively represent the spike emission intervals of the corresponding spike streams.

By solving Eq. \ref{eq:s2}, the additional brightness of the scene due to the equivalent dark current can be calculated:
\begin{equation}
    L_d = \frac{L_1\,T_1}{T_d-T_1}.
    \label{eq:s3}
\end{equation}

Another spike streams with brightness $L_2$ needs to be captured to nullify the unknown photoelectric conversion constant $\alpha$:
\begin{equation}
	C\Delta V = \alpha (L_2 + L_d)T_2.
    \label{eq:s4}
\end{equation}

Due to mismatches in capacitance and voltage between pixel circuits, different pixels exhibit varying responses to scene brightness, leading to fixed pattern noise. The corresponding error matrix can be defined as follows:
\begin{equation}
    R(x,y) = \frac{(C+\delta C(x, y))\,(\Delta V+\delta V(x, y))}{C \Delta V}.
\end{equation}

Substituting Eq. \ref{eq:s3} and Eq. \ref{eq:s4} into the above equation yields a specific value for $R(x,y)$:
\begin{equation}
    R(x,y)=\frac{(L_2+L_d(x_m, y_m))T_2(x_m,y_m)}{(L_2+L_d(x, y))T_2}.
    \label{eq7}
\end{equation}


According to Eq. \ref{eq7}, the threshold variation of spiking neurons can be simulated based on the real-world spike distribution.

\section{Additional Quantitative Results}
The detailed quantitative results on six synthetic scenarios are shown in Table~\ref{tab:s2} and Table ~\ref{tab:s3}.
Table~\ref{tab:s2} illustrates the superior performance of SpikeNeRF across all synthetic scenarios. This observation suggests that our method excels in learning a more precise 3D representation of the scene within the proposed framework.

In Table~\ref{tab:s3}, we present the outcomes for the synthetic spike dataset under various light intensities. These diverse light conditions are achieved by adjusting the intensity parameter in the spike simulator, with settings for low (16), medium (32), and strong (64) illumination. The spike numbers, representing the count of generated spike data for a view, are detailed in the table. A higher spike number indicates a stronger light intensity.
The results consistently reveal the superior performance of our final model compared to other configurations.




\begin{table*}[t]
\renewcommand\arraystretch{1}
\small
\setlength{\tabcolsep}{2.6mm}
\centering
\caption{Quantitative evaluation of different light intensities on synthetic dataset.}
\begin{tabular}{cccccc}
\toprule[1pt]
\multirow{2}{*}{Method}&\multirow{2}{*}{Loss}&\multicolumn{4}{c}{Light intensity (16)} \\
\cmidrule(r){3-6} 

&& PSNR $\uparrow$ & SSIM $\uparrow$ & LPIPS & $\#$SpikeNum \\
\hline
TFP(32)+NeRF&MSE& 15.27/15.57& 0.154/0.785& 0.588/0.065&\multirow{8}{*}{268,185}  \\
TFP(256)+NeRF&MSE& 14.44/16.41&0.127/0.794& 0.652/0.088  \\ 
TFI+NeRF&MSE& 13.77/16.63& 0.120/0.793& 0.670/0.093\\
Spk2img+NeRF&MSE& 13.42/14.05& 0.066/0.726& 0.724/0.129\\ \cmidrule(r){1-5}

\multirow{4}{*}{Ours}& $\mathcal{L}_i$$^{*}$& 13.77/16.63& 0.120/0.793&0.670/0.093\\
& $\mathcal{L}_i$& 18.55/\textbf{19.76}& 0.237/0.878& 0.527/0.051 \\ 

&$\mathcal{L}_s$+$\mathcal{L}_i$& 18.50/19.59& 0.237/0.876& 0.524/0.051\\ 
&$\mathcal{L}_s$& \textbf{18.72}/19.38& \textbf{0.251}/\textbf{0.880}& \textbf{0.517}/\textbf{0.050}\\

\toprule[1pt]
\multirow{2}{*}{Method}&\multirow{2}{*}{Loss}&\multicolumn{4}{c}{Light intensity (32)} \\
\cmidrule(r){3-6} 

&& PSNR $\uparrow$ & SSIM $\uparrow$ & LPIPS  & $\#$SpikeNum\\
\hline
TFP(32)+NeRF&MSE& 17.81/18.55& 0.191/0.827& 0.562/0.077&\multirow{8}{*}{414,475} \\
TFP(256)+NeRF&MSE& 15.82/18.68& 0.170/0.835& 0.616/0.077  \\ 
TFI+NeRF&MSE& 15.07/18.53& 0.156/0.825& 0.634/0.083\\
Spk2img+NeRF&MSE& 14.59/15.39& 0.109/0.768& 0.651/0.093\\ \cmidrule(r){1-5}

\multirow{4}{*}{Ours}& $\mathcal{L}_i$$^{*}$& 15.07/18.53& 0.156/0.825& 0.634/0.083\\
& $\mathcal{L}_i$& 21.21/22.81& 0.272/0.890& 0.492/0.055 \\ 

&$\mathcal{L}_s$+$\mathcal{L}_i$& 21.76/23.59& 0.279/0.913& 0.486/0.053\\ 
&$\mathcal{L}_s$& \textbf{22.05}/\textbf{23.66}& \textbf{0.300}/\textbf{0.926}& \textbf{0.477}/\textbf{0.050}\\

\toprule[1pt]
\multirow{2}{*}{Method}&\multirow{2}{*}{Loss}&\multicolumn{4}{c}{Light intensity (64)} \\
\cmidrule(r){3-6} 

&& PSNR $\uparrow$ & SSIM $\uparrow$ & LPIPS & $\#$SpikeNum\\
\hline
TFP(32)+NeRF&MSE& 20.63/21.86& 0.239/0.872& 0.528/0.067 &\multirow{8}{*}{720,032}  \\
TFP(256)+NeRF&MSE& 17.28/21.27& 0.189/0.852& 0.596/0.081  \\ 
TFI+NeRF&MSE& 16.16/20.56& 0.187/0.852& 0.602/0.077\\
Spk2img+NeRF&MSE& 16.37/17.73& 0.169/0.837& 0.595/0.073\\ \cmidrule(r){1-5}

\multirow{4}{*}{Ours}& $\mathcal{L}_i$$^{*}$ & 16.16/20.56& 0.187/0.852& 0.602/0.077\\
& $\mathcal{L}_i$& 21.28/21.98& 0.334/0.874& 0.463/0.068 \\ 

&$\mathcal{L}_s$+$\mathcal{L}_i$& \textbf{24.09}/\textbf{24.85}& 0.382/0.920& 0.442/0.051\\ 
&$\mathcal{L}_s$& 23.89/24.46& \textbf{0.411}/\textbf{0.929}& \textbf{0.428}/\textbf{0.049}\\ \toprule[1pt]

\end{tabular}
\vspace{50mm}
\label{tab:s3}
\end{table*}

\section{Additional Qualitative Results}
\begin{figure*}[t]
  \centering
   \includegraphics[width=1\linewidth]{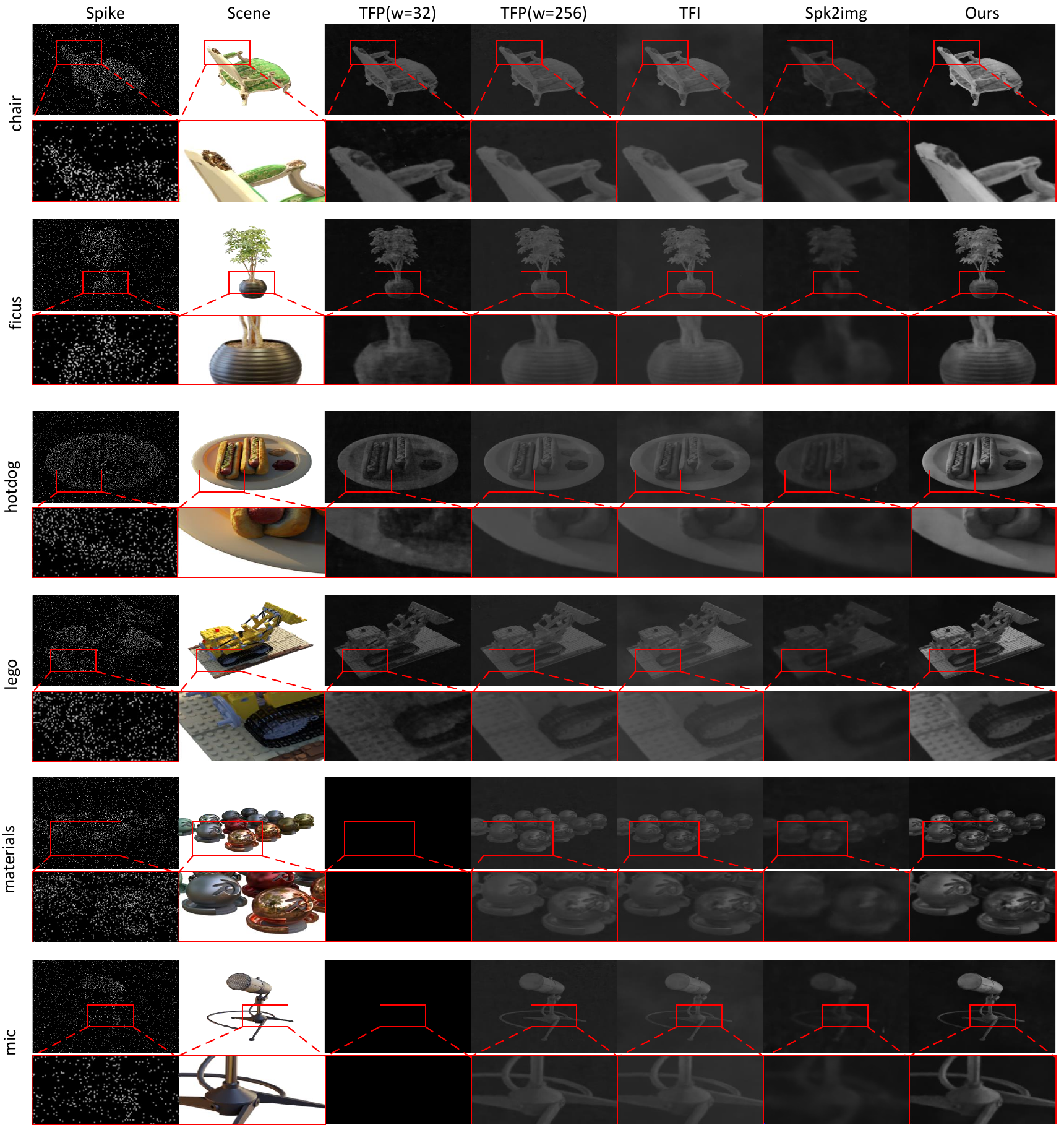}

   \caption{Quantitative results on synthetic spike data.}
   \label{fig:s3}
\end{figure*}

\begin{figure*}[t]
  \centering
   \includegraphics[width=1\linewidth]{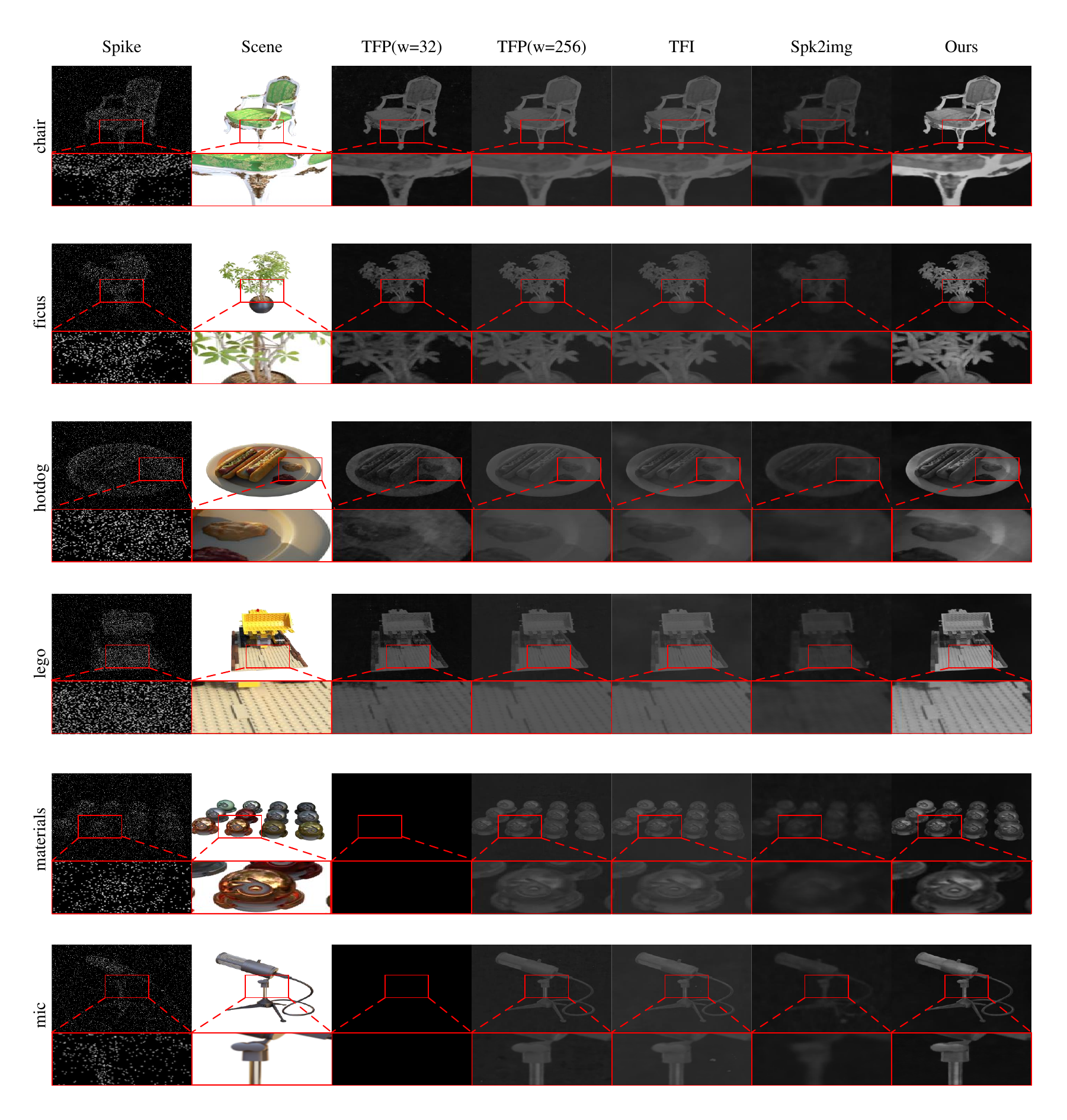}

   \caption{Quantitative results on synthetic spike data.}
   \label{fig:s4}
\end{figure*}

\begin{figure*}[t]
  \centering
   \includegraphics[width=1\linewidth]{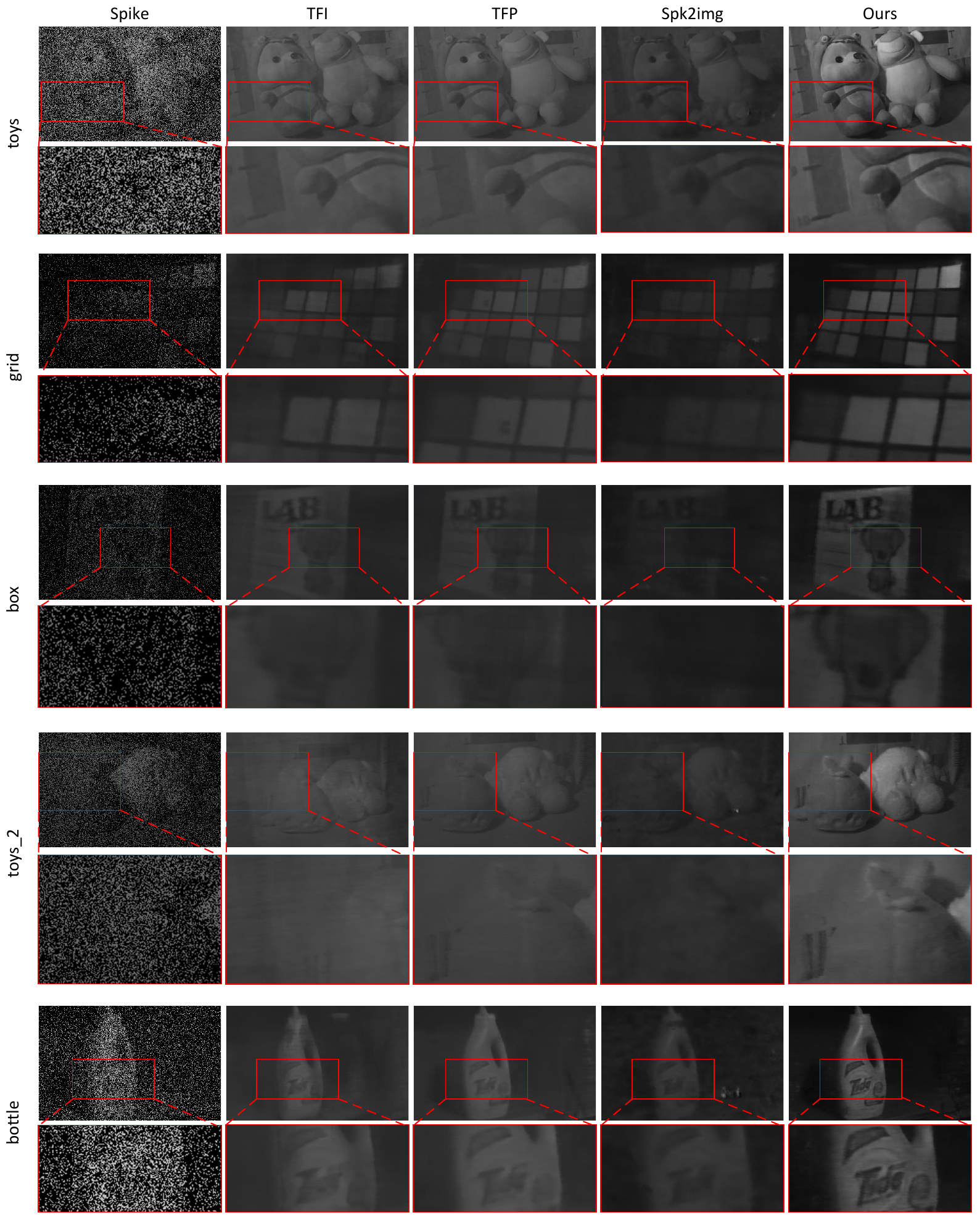}

   \caption{Quantitative results on real-world spike data.}
   \label{fig:s5}
\end{figure*}

The qualitative results on synthetic scenarios and real-world spike data are shown in Fig.~\ref{fig:s3}, Fig.~\ref{fig:s4}, and Fig.~\ref{fig:s5}.

In Fig.~\ref{fig:s3} and Fig.~\ref{fig:s4}, our SpikeNeRF effectively leverages the inherent relationship between spike streams and scenes to learn a sharp NeRF. Consequently, our results remain resilient to the noise inherent in spike data, yielding impressive results in novel view synthesis.

Fig.~\ref{fig:s5} showcases the results obtained from five real-world spike sequences. When confronted with spike data noise, other methods exhibit limitations, introducing more noise into their output. In contrast, our SpikeNeRF excels in predicting accurate details and light intensities compared to alternative methods.

\section{Additional Ablation Results}


\noindent\textbf{Comparison to finetuned/retrained Spk2imgNet.}
We conduct separate retraining and fine-tuning of Spk2imgNet on the dataset provided by RSIR, and the results are presented in Table~\ref{tab:s4}. 
We show the results focused solely on measurements within the object region. 
Since the Spk2imgNet network is primarily designed for high-intensity lighting conditions, and the lighting situations in the simulated and real-world scenarios addressed in this paper are more complex, training Spk2imgNet with a noisy dataset makes it more challenging for the network to learn the correspondence between spikes and light intensity, resulting in no improvement in performance.
RSIR processes spike data using a cyclic iterative approach, where, in the original study, spikes of length 32 or 64 are inputted in each iteration, and the loop performs optimally after 4-8 iterations. Therefore, we design three configurations for comparison, as shown in the table, where `w' represents the length of each input spike, and `c' represents the number of iterations. According to SSIM and LPIPS metrics, it can be observed that among these configurations, RSIR(w=32, c=8) performs best.

\noindent\textbf{Ablation of the nonuniformity matrix.}
We conduct an ablation experiment of nonuniformity matrix $R$ in Table~\ref{tab:s5}. 
In Table 2 of our paper, we also compare our proposed loss $\mathcal{L}_s$ with $\mathcal{L}_i$, $\mathcal{L}_s$+$\mathcal{L}_i$, and $\mathcal{L}_i^*$.


\noindent\textbf{Different sequence lengths.}
In Table~\ref{tab:s6}, we conduct experiments using sequences of different lengths.
We observed that PSNR might be more significantly influenced by the contrast, with the highest results achieved using w=64 under light intensity 16.
SSIM and LPIPS metrics focus more on measuring structural and perceptual information, respectively. As the length of the spike sequence increases, the SSIM and LPIPS metrics improve under different light intensities.
SpikeNeRF can converge under low illumination and with a short sequence length.

\begin{table*}[htbp]
\small
\tabcolsep=1.32mm
\centering
\caption{Compasrison to Spk2imgNet and RSIR.}
\begin{tabular}{ccccccccccccc}
\toprule[1pt]
\multirow{2}{*}{Method}&\multicolumn{3}{c}{Light intensity (16)} &\multicolumn{3}{c}{Light intensity (32)} &\multicolumn{3}{c}{Light intensity (64)}\\
\cmidrule(r){2-4} \cmidrule(r){5-7} \cmidrule(r){8-10}

& PSNR  & SSIM  & LPIPS & PSNR  & SSIM  & LPIPS & PSNR  & SSIM  & LPIPS \\
\hline

  

Spk2imgNet-finetuned& 13.38&0.712	&0.121&	14.78&	0.749	&0.100&	16.92	&0.793	&  0.091
  \\
Spk2imgNet-retrained& 13.60	&0.718&	0.114&	14.98&	0.756	&0.095&	17.16	&0.806	&0.081
 \\ 
RSIR(w=32,c=8)& 14.58 &0.754	&\underline{0.087}&	\underline{16.49}&	\underline{0.906}	&\underline{0.072}&	\underline{19.58}	&\underline{0.848}	&\underline{0.071}
\\
RSIR(w=64,c=4)& \underline{14.71}	&\underline{0.757}	&0.088	&16.42	&0.794	&0.083&	19.51	&0.846	&\underline{0.071}
\\ 
RSIR(w=256,c=1)& 14.69&	\underline{0.757}	&0.088&	16.46	&0.803&	0.076&	19.48&	0.846&	\underline{0.071}
\\ \hline
Ours& \textbf{19.38}& \textbf{0.880}& \textbf{0.050}& \textbf{23.66}& \textbf{0.926}& \textbf{0.050}& \textbf{24.46}& \textbf{0.929}& \textbf{0.049}\\ \toprule[1pt]
\end{tabular}
\label{tab:s4}
\end{table*}

\begin{table*}[t]
\renewcommand\arraystretch{1}
\small
\tabcolsep=1.32mm
\centering
\caption{Ablation of the nonuniformity matrix.}
\begin{tabular}{ccccccccccccc}
\toprule[1pt]
\multirow{2}{*}{Method}&\multicolumn{3}{c}{Light intensity (16)} &\multicolumn{3}{c}{Light intensity (32)} &\multicolumn{3}{c}{Light intensity (64)}\\
\cmidrule(r){2-4} \cmidrule(r){5-7} \cmidrule(r){8-10}

& PSNR  & SSIM  & LPIPS & PSNR  & SSIM  & LPIPS & PSNR  & SSIM  & LPIPS \\
\hline
TFP&16.41& 0.794 &0.088 &18.68 &0.835& 0.077 &21.27 &0.852& 0.081\\

TFI&16.63& 0.793 &0.093& 18.53& 0.825& 0.083 &20.56& 0.852& 0.077\\

Ours(w/o $R$, w=256)&	16.23	&0.819	&0.052	&19.66&	0.886	&0.048&	23.67	&0.910	&0.058\\

Ours(w=256)	& \textbf{19.38}&	\textbf{0.880}	&\textbf{0.050}	&\textbf{23.66} &	\textbf{0.926}	&\textbf{0.050} &	\textbf{24.46}	&\textbf{0.929}	&\textbf{0.049}

\\
\toprule[1pt]
\end{tabular}
\label{tab:s5}
\end{table*}

\begin{table*}[t]
\renewcommand\arraystretch{1}
\small
\tabcolsep=1.32mm
\centering
\caption{Comparisons on different sequence lengths.}
\begin{tabular}{ccccccccccccc}
\toprule[1pt]
\multirow{2}{*}{Method}&\multicolumn{3}{c}{Light intensity (16)} &\multicolumn{3}{c}{Light intensity (32)} &\multicolumn{3}{c}{Light intensity (64)}\\
\cmidrule(r){2-4} \cmidrule(r){5-7} \cmidrule(r){8-10}

& PSNR  & SSIM  & LPIPS & PSNR  & SSIM  & LPIPS & PSNR  & SSIM  & LPIPS \\
\hline

  
Ours(w=32)&	16.94	&0.767&	0.145	&25.29	&0.894	&0.059&	23.74	&0.909	&0.054 \\

Ours(w=64)&	\textbf{20.59}	&0.876&	0.053&	25.92	&0.918&	0.053&	23.18&	0.915&	0.056 \\

Ours(w=128)	&19.77&	0.879	&0.052&	\textbf{25.93}	&0.925&	\textbf{0.050}	&23.85&	0.922	&0.052\\ 

Ours(w=256)	&19.38&	\textbf{0.880}&	\textbf{0.050}	&23.66&	\textbf{0.926}	&\textbf{0.050} &	\textbf{24.46}	&\textbf{0.929}&	\textbf{0.049}

\\
\toprule[1pt]
\end{tabular}
\label{tab:s6}
\end{table*}


\section{Supplementary Video}
We provide a supplementary video to show the video results.
For synthetic scenes, we show the results of TFI+NeRF, TFP+NeRF, Spk2img+NeRF and our SpikeNeRF.
For the real scenes, we show the comparison of the results of all mentioned methods in the toy and toy$\_$2 scenes.
It is obvious that the results of our SpikeNeRF have less noise, better contrast, and sharper object texture details compared to other methods on both synthetic scenes and real scenes.


\end{document}